# Parallelizing Probabilistic Inference
# Some Early Explorations

Bruce D'Ambrosio
Tony Fountain
Zhaoyu Li
Department of Computer Science
Oregon State University
Corvallis, OR 97331-3202
dambrosi@cs.orst.edu

## Abstract

We report on an experimental investigation into opportunities for parallelism in belief-net inference. Specifically, we report on a study performed of the available parallelism, on hypercube style machines, of a set of randomly generated belief nets, using factoring (SPI) style inference algorithms. Our results indicate that substantial speedup is available, but that it is available only through parallelization of individual conformal product operations, and depends critically on finding an appropriate factoring. We find negligible opportunity for parallelism at the topological, or clustering tree, level.

## 1 Introduction

Probabilistic inference in belief nets is a computationally intensive process [2]. One way to reduce the time required might be to use parallel hardware, but little has been reported to date on attempts to parallelize belief-net inference algorithms. In this paper we report the results of studies on speedup opportunities for one class of inference algorithms, those based on a direct symbolic-reasoning attack on the underlying factoring problem [10], for one class of practical parallel machines, hypercubes[1]. We find that good speedup is available for large problems, but that alternative factorings of equivalent single processor complexity can have widely varying performance on parallel machines due to varying communication costs. We present a simple factoring heuristic which performs surprisingly well, and analyze the reasons for its good performance.

## 2 Background

We begin with a review of our basic algebraic approach to probabilistic inference and of opportunities for parallelism within this approach.

### 2.1 Factoring Approaches to Probabilistic Inference

Symbolic-algebraic approaches to belief-net inference were first described in [3], [10], although see [11] for closely related work. The essential element of these approaches is the construction of an evaluation tree specifying how to combine the belief-net distributions relevant to a query. The operations performed at each node of the tree are a conformal product of the results of the two immediate child nodes and a summation over variables not needed higher in the tree. There are many correct evaluation trees corresponding to a typical query, and the problem of finding an optimal (minimal computational complexity) tree is hard. In this paper we consider alternate heuristics for evaluation tree construction and examine the impact of the alternative evaluation trees on speedup available. One heuristic we consider is SPI [4]. SPI uses the global graph structure as a constraint on the evaluation tree for a query (by forming a "partition tree" which guides evaluation tree construction). Preliminary studies [5] revealed that SPI evaluation trees do not parallelize well. We then tested two other heuristics in order to determine whether the poor performance of SPI trees was due to an inherent complexity of probabilistic inference, or was an artifact of the particular evaluation trees constructed by SPI.

The first heuristic we tested was a greedy heuristic we call "set-factoring(s)" [8] that constructs evaluation trees without reference to graph structure, simply attempting at each stage to find the lowest cost conformal product it can perform. For this heuristic "cost" is measured as time and space complexity of a conformal product. Experimental tests have shown that set-factoring performs quite well at sequential probabilistic inference, typically outperforming the current implementation of SPI. We then tested a second "improved" heuristic "set-factoring(c)[2], similar to set-factoring but with the cost heuristic extended to include communication costs. Since the reason for low speedup available on SPI trees was high communication cost, we naturally expected that the second heuristic would generate more parallelizable evaluation trees than the original set-factoring algorithm. We were pleased to discover that the evaluation trees generated by set-factoring(s) parallelize quite well, but surprised to learn that the evaluation trees generated by the "improved" heuristic, set-factoring(c) provided no further improvement.

---

[1]Acknowledgement: This research supported by NSF 91-00530 and a grant from OACIS.

[2]"C" for with communication cost.

## 2.2 Sources of Parallelism

For this investigation we considered two sources of parallelism in query evaluation. One source of parallelism in the query evaluation process is at the evaluation-tree level. Since each non-leaf node in an evaluation tree corresponds to a conformal product operation, and since distributions occur at most once in an evaluation tree, the conformal product operations corresponding to nodes in disjoint subtrees can be performed in parallel. A lower bound on the computational time for parallel evaluation using this approach is the length of time it would take to evaluate the longest path in the evaluation tree, where path length is measured in terms of the number of multiplies required for the conformal products in the path. Using the same approach as was used in calculating the lower bound, we can calculate an upper bound by summing the number of multiplies for each conformal product in the evaluation tree. It is clear that there will not be much speedup unless the evaluation trees are relatively bushy.

A second source of parallelism in query processing is at the conformal-product level. That is, each conformal product operation can itself be parallelized. Our approach to conformal product parallelism is based on calculating subsets of the entries in the result distribution in parallel. For distributed-memory architectures, such an approach requires splitting the input distributions in such a manner that each processor gets the input values necessary for computing its respective portion of the result distribution. Two important points to consider in this approach are: (1) splitting the input distributions on variables that occur in both input distributions minimizes the total amount of data communication required to compute the conformal product, and (2) splitting the input distributions on variables other than those in the result distribution requires either the ability to perform concurrent writes with summing or a separate processing step in which the values that make up the result distribution entries are summed together. This latter occurs because a split on variables other than result variables causes the result distribution subsets computed at separate processors to overlap. To simplify our analysis, we only considered the parallelism available from splitting on subsets of the variables in the result distribution.

## 3 Models

The parallel model of computation used for this investigation was based on a broadcast-compute-aggregate (BCA) model of conformal product evaluation. Under this model a distinguished processor is responsible for partitioning each conformal product into subtasks, which are distributed to the other processors for evaluation. This same distinguished processor collects the results from the subtasks and assembles them into the result distribution. This model is communication-intensive, since all data originates and returns to a single processor. An alternative to the BCA approach is the distributed-net model (Dist-net), in which the entire belief net, including all intermediate results, is stored on each processor. Compared to the BCA model, the Dist-net model trades memory for communication costs. For this paper we concentrate our discussion on the BCA model. We do, however, present some preliminary results at the end of this paper comparing the Dist-net and BCA models.

We use two-level models of sequential and parallel inference in this paper. First, we use a low-level model of individual conformal-product evaluation. Second, we use a higher-level model of query-tree evaluation. Each of the models was parameterized to allow experimentation with various aspects of the computational environment including task grainsizes, number of processors, and communication costs. Some details of the actual models have been omitted from what follows for brevity. For complete detail see[5].

### 3.1 Sequential Model

To provide a basis against which to measure the performance of the parallel algorithms, we developed the following model to estimate the running time of the sequential algorithm.

Sequential conformal product model: $T_s(c) = \alpha M(c)$ where $M(c)$ is the number of multiplies in conformal product $c$ and $\alpha$ is a constant scaling factor that accounts for factors other than multiplies in the conformal product operation, most notably indexing and additions. $M(c)$ is calculated by $\prod_{v \in V} n(v)$, where $V$ is the union of the variables occurring in the two input distributions. To simplify analysis we restricted the value space of the variables in the test belief nets to 2 values each. Thus in the following sections $M(c) = 2^n$ where n is the number of unique variables in the 2 input distributions.

Given the model for the sequential running time for a single conformal product, the estimated sequential running time for the evaluation of an entire query is the sum of the time required for each of its conformal products.

Sequential model for a query: $T_s(q) = \sum_{c \in q} T_s(c)$

### 3.2  Conformal Product Parallel Model

The following model was developed to estimate the running time of a parallel algorithm.

Parallel conformal product model: $T_p(c) = P + S + W + C$

Where
$P$ is the cost for process initialization,
$S$ is the cost for setting the problem up,
$W$ is the cost for the work done at each processor, and
$C$ is the cost for communication.

The value which was used for $W$ was $\alpha G$, where $G$ is the computational grainsize specified as the number of floating point multiplies per process. $\alpha$ is the same constant scaling factor that was used in the sequential model.

The following measures were calculated for each conformal product:

$T_s$ = Time for sequential
$T_p$ = Time for parallel
$N_u$ = Number of processors used

The values for $G$ and $N_u$ were determined as follows:

1. A minimum grainsize, $G_{min}$, and a maximum number of processors, $N_a$, were specified.

2. Given a particular conformal product to compute, the actual $G$ and $N_u$ values were calculated so that $N_u$ was maximized under the constraints $N_u \leq N_a$, $G \geq G_{min}$, and $N_u \leq 2^v$, where $v$ is the number of variables in the result distribution. In other words, $N_u$ and $G$ were chosen such that as many as possible of the available processors were used as long as there was enough work for each processor to perform as specified by the minimum grainsize, and there was enough parallelism in the problem to support the desired partitioning.

#### 3.2.1  Distributed-Memory Communication Model

The distributed-memory model includes a specific model of communication for a cube architecture. This model assumes that there is no overlap between the conformal product calculations and the communication between processors. This model also assumes that the data to be sent to processors is arranged into buffers, one buffer for each process.

Total communication cost: $C = C_d + C_r + B$
where
$C_d$ is the communication cost for distributing the data,
$C_r$ is the cost of returning the data, and
$B$ is the cost for building the buffers of data to be distributed.

The data transfer cost calculations were based on the log spanning tree, or broadcast, communication model for hypercube [6].

Distribution communication cost: $C_d = (D_{max} * C_{st}) + (B_d * ((N_u - 1) * C_b))$
where
$D_{max}$ is the maximum dimension of the cube which is used,
$C_{st}$ is the communication startup time,
$B_d$ is the number of bytes sent to each processor, and
$C_b$ is the communication cost per byte, per link.

The following formula was used to calculate $B_d$, the number of bytes sent to each processor:

$B_d = B_{total} / N_u$
where $B_{total}$ is the total number of bytes needed to compute the conformal product[3].

The return communication cost is calculated in the same way as distribution communication costs:
Return communication cost: $C_r = (D_{max} * C_{st}) + (B_r * ((N_u - 1) * C_b))$
where
$B_r$ is the number of bytes returned from each processor and is calculated by
$B_r = B_{result} / N_u$
where $B_{result}$ is the total number of bytes in the result distribution. Since there are $2^{|result-dist-vars|}$ entries in the result distribution, $B_{result} = 4 * 2^{|result-dist-vars|}$.

### 3.3  Parallel Model of Query Evaluation

As explained earlier, query evaluation consists of repeated conformal product operations. Since we were interested in the performance of a parallel inference algorithms on the task of query evaluation we constructed a model to predict this performance from the models for conformal product operations. The parallel query model is analogous to the sequential query model. The running time for parallel query evaluation is simply the sum of the running times of its conformal products.

Parallel model for a query: $T_p(q) = \sum_{c \in q} T_p(c)$

$N_u(q)$, the number of processors used in evaluating query $q$, is simply the maximum of the number used by any of the conformal products in $q$. Given $T_p(q)$, $T_s(q)$ and $N_u(q)$ the speedup, cost, and efficiency for a query can be calculated according to formulas given in [1].

Speedup: $S(q) = T_s(q) / T_p(q)$

Cost: $C(q) = T_p(q) * N_u(q)$

Efficiency: $E(q) = T_s(q) / C(q) = S(q) / N_u(q)$

### 3.4  Evaluation Tree Parallelism Models

For evaluation tree parallelism, we only computed a lower bound on running time. As mentioned earlier, a lower bound on the running time of an algorithm exploiting evaluation tree parallelism can be calculated by summing the times required to perform each of the conformal products in the longest path of the evaluation tree.

---

[3]$B_{total}$ is, in general, greater than the sum of the sizes of the two input distributions. As usual, see [5] for details.

Lower bound on $T_p(q) = \sum_{c \in L_p} T_s(c)$ where $Lp$ is the longest path in the evaluation tree as measured by the amount of time it takes to compute the conformal products in the path.

## 4  Method

We used an experimental method to explore opportunities for parallelism in belief net inference. Specifically, we generated a set of belief nets (and observations and a query for each net), used each of the three methods previously described to generate an evaluation tree for the query, then used our cost models to compute the cost of performing the required numeric computations sequentially, using BCA, and using Dist-net.

The belief nets were generated using J. Suermondt's random-net generator under the following constraints: for each belief net, the number of nodes was randomly chosen between 10 and 100; the average arcs per node were between 1 and 5; and the number of observations was between 1 and 20. The query node was chosen at random from the nodes in the net that were not observed. All variables had two values.

Table 1 provides the description of the eight random nets we measured. Values in the nodes column show the total number of nodes in each belief net; arcs shows the average number of incoming arcs per node; obs gives the number of observations posted to the net; CPs is the number of conformal products in the query[4]; and finally seq-time is the best sequential time across all three algorithms(in microseconds, based on our sequential cost model)[5]. We use this time as the sequential time for computing absolute speedup.

Since we used a model rather than an actual parallel implementation, we had to make assumptions about the number of processors available, processor speeds, communication costs, and so on. We assumed a maximum of 1024 processors, and chose a minimum task grainsize of 256. We set $\alpha$, the scaling factor for multiplication, at 45 microseconds. $P$, the cost for process initialization, was taken to be 0. $S$, the cost for setting the problem up, was 0. $C_{st}$, the communication start-up time, was 230 microseconds. $C_b$, the communication cost per byte, was 0.5 micro seconds per byte per link. $B$, the cost for building the buffers of data to be distributed, was 0. These values are believed representative of actual costs on an Intel IPSC-2, and are based on discussions with the parallel algorithms and languages groups at OSU. Further, the relative values of these numbers seem to be valid for announced and foreseeable hypercube-style machines. Since our speedup measurements are dependent on the relative values, rather than the absolute values, we expect that our results are applicable to most machines in this architecture class. For further discussion see [5].

## 5  Results

We begin with the basic measurements using the BCA model for each of our eight nets for each of the three algorithms. For each algorithm, we measured:

- **dm**, the maximum dimensionality of any conformal product in the evaluation tree,
- **md**, the value of $max(d1, d2, r)$, where $d1$, $d2$ and $r$ are the size (number of variables) of two input distributions and the result in the largest conformal product of a query.
- **dd**, $(dm - md)/dm$.
- **cm-cst**, the total communication cost for the query,
- **cp-cst**, the total computation cost for the query,
- **cp/cm**, the ratio of the computation cost to the communication cost,
- **ttl-cst**, the total execution time,
- **r-spdp**, relative speedup: the ratio of uniprocessor time with multi-processor time for the same evaluation tree, and
- **a-spdp**, the absolute speedup, the ratio of uniprocessor time for the best evaluation tree to multi-processor time for this tree.

Tables 2, 3, and 4 contains the corresponding measured values for the basic set-factoring heuristic. All queries were able to make use of all 1024 processors at least part of the time except query number 4 in table 3.

Table 5 presents data comparing communication and memory requirements of set-factoring with communication cost heuristic under the BCA model and the dist-net model. The information in this table is as follows:

- **BCA cm**: communication cost for the BCA model.
- **Dist cm**: communication cost for the dist-net model.
- **BCA mem**: memory size used by the BCA model, estimated as the total number of bytes communicated to and from each processor.

---
[4] The number of CPs is equal to the total number of factors minus one. Since the SPI version used in [5] may use more than the minimal required set of nodes to compute a query, the number of CPs for SPI may different.

[5] Throughout the rest of this paper, all communication cost, computation cost and total cost in any table are in microseconds.

| # | nodes | arcs | obs | CPs | seq-time |
|---|---|---|---|---|---|
| 1 | 70 | 4.2 | 6 | 51 | 5.41+9 |
| 2 | 68 | 3.5 | 10 | 48 | 1.03+8 |
| 3 | 86 | 4.3 | 13 | 64 | 1.60+12 |
| 4 | 50 | 4 | 16 | 41 | 2.19+7 |
| 5 | 65 | 4.9 | 7 | 50 | 6.08+11 |
| 6 | 89 | 4.2 | 3 | 62 | 2.08+14 |
| 7 | 83 | 4.2 | 9 | 54 | 3.75+10 |
| 8 | 84 | 3.7 | 6 | 46 | 4.04+7 |

Table 1: Random net descriptions.

| # | dm | md | dd | cm-cst | cp-cst | cp/cm | ttl-cst | r-spdp | a-spdp |
|---|---|---|---|---|---|---|---|---|---|
| 1 | 29 | 28 | 0.03 | 4.08+9 | 8.39+7 | 0.02 | 4.17+9 | 20.0 | 1.3 |
| 2 | 26 | 25 | 0.04 | 6.20+8 | 1.32+7 | 0.02 | 6.34+8 | 6.3 | 0.2 |
| 3 | 34 | 33 | 0.03 | 2.48+11 | 1.37+9 | 0.02 | 2.53+11 | 20.1 | 6.3 |
| 4 | 20 | 19 | 0.05 | 1.77+7 | 3.99+5 | 0.02 | 1.81+7 | 16.4 | 1.2 |
| 5 | 34 | 34 | 0.00 | 2.04+11 | 3.68+9 | 0.02 | 2.08+11 | 18.2 | 2.9 |
| 6 | 46 | 45 | 0.02 | 4.52+14 | 9.80+12 | 0.02 | 4.6+14 | 21.8 | 0.5 |
| 7 | 32 | 31 | 0.03 | 2.27+10 | 4.99+8 | 0.02 | 2.32+10 | 22.0 | 1.6 |
| 8 | 22 | 22 | 0.00 | 6.30+7 | 1.38+6 | 0.02 | 6.40+7 | 18.7 | 0.6 |

Table 2: Test results for the SPI algorithm.

| # | dm | md | dd | cm-cst | cp-cst | cp/cm | ttl-cst | r-spdp | a-spdp |
|---|---|---|---|---|---|---|---|---|---|
| 1 | 29 | 25 | 0.14 | 2.2+8 | 4.8+7 | 0.22 | 2.7+8 | 159.6 | 20.0 |
| 2 | 20 | 17 | 0.15 | 1.04+6 | 4.76+5 | 0.46 | 1.52+6 | 67.9 | 67.9 |
| 3 | 36 | 31 | 0.14 | 9.2+9 | 3.6+9 | 0.39 | 1.2+10 | 256.9 | 133.3 |
| 4 | 18 | 15 | 0.17 | 3.39+5 | 2.76+5 | 0.81 | 6.15+5 | 35.5 | 35.5 |
| 5 | 33 | 29 | 0.12 | 2.98+9 | 6.40+8 | 0.22 | 3.62+9 | 168.0 | 168.0 |
| 6 | 42 | 35 | 0.17 | 1.65+11 | 2.03+11 | 1.23 | 3.68+11 | 563.9 | 563.9 |
| 7 | 30 | 24 | 0.20 | 1.46+8 | 7.86+7 | 0.54 | 2.25+8 | 249.4 | 169.8 |
| 8 | 23 | 18 | 0.22 | 1.64+6 | 1.34+6 | 0.82 | 2.98+6 | 148.3 | 13.5 |

Table 3: Test results for set-factoring.

| # | dm | md | dd | cm-cst | cp-cst | cp/cm | ttl-cst | r-spdp | a-spdp |
|---|---|---|---|---|---|---|---|---|---|
| 1 | 26 | 22 | 0.15 | 3.11+7 | 1.73+7 | 0.56 | 4.84+7 | 111.8 | 111.8 |
| 2 | 22 | 17 | 0.23 | 1.48+6 | 4.45+5 | 0.30 | 1.93+6 | 123.95 | 53.5 |
| 3 | 35 | 28 | 0.20 | 1.70+9 | 1.50+9 | 0.88 | 3.30+9 | 479.8 | 479.8 |
| 4 | 19 | 15 | 0.21 | 3.16+5 | 2.35+5 | 0.74 | 5.51+5 | 52.2 | 39.7 |
| 5 | 35 | 30 | 0.14 | 5.98+9 | 2.39+9 | 0.40 | 8.37+9 | 280.9 | 72.6 |
| 6 | 42 | 35 | 0.17 | 1.51+11 | 2.42+11 | 1.60 | 3.93+11 | 567.8 | 528.3 |
| 7 | 29 | 25 | 0.14 | 1.78+8 | 8.40+7 | 0.47 | 2.62+8 | 142.0 | 142.0 |
| 8 | 19 | 15 | 0.21 | 3.64+5 | 2.43+5 | 0.67 | 6.08+5 | 66.5 | 66.5 |

Table 4: Test results for set-factoring(c).

- **Dist mem**: memory size used in the dist-net model, estimated as the size of data in the two input distributions plus the result size, for the the largest conformal product.
- **memory**: Memory size used by BCA model ignoring the final conformal product, see discussion.
- **mem/Dist-mem**: Ratio of memory to Dist-mem.

Finally, table 6 shows information relevant to evaluation-tree parallelism for the set-factoring with communication cost algorithm:

- **para-cp**: the number of conformal products that can be parallelized (using grainsize 256).
- **%cp**: the number of conformal products parallelizable as a percentage of the total number of conformal products.
- **lp-cp**: the number of conformal products in the longest path.
- **lp-speedup**: the speedup using evaluation-tree parallelism as well as conformal product parallelism. In table 4, ttl-cst is the total execution time for a query without considering any evaluation-tree parallelism. We estimate the further speedup available through evaluation tree parallelism by considering only the highest cost leaf-to-root path in the evaluation tree[6].
- **lp-%-cp**: the number of conformal products in the longest path as a fraction of the total number of conformal products.
- **%-time**: the ratio of the total cost of evaluation-tree parallelism plus conformal product parallelism with conformal product parallelism alone.

## 6  Discussion

**Parallel computation of belief net inferences is feasible**  From table 3 and table 4 we can see that good absolute speedup is available for queries with high dimensionality. The poor results in our earlier attempt to parallelize SPI evaluation-trees apparently do not accurately reflect the parallelism available in the underlying computation. Some queries show relatively little speedup, for example nets 2, 4 and 8. These queries have lower maximum dimensionalities, and simply are too small to effectively parallelize (We will see later that the factorings for those queries are as good as the factorings for the other queries in our experiment.).

**Characteristics of a Parallelizable Evaluation Tree**  When constructing an evaluation tree for sequential computation, only the maximum dimensionality $max\_dim$ of the tree is important. However, in constructing evaluation trees for parallel computation there are three factors to consider. First, we must consider the maximum dimensionality, as for sequential computation. Second, we must consider the sizes of the two input distributions and the result distribution (for the largest conformal product). As we shall discuss below, this will affect the communication costs. Third, we must consider the degree to which the evaluation tree is balanced. This will determine the available evaluation-tree parallelism.

Computation cost for a conformal product is exponential in $max\_dim$. Under optimum conditions (grain size, result variables available for splitting, etc) we can reduce this cost by a factor of $n$ by distributing the computation over $n$ processors. Nonetheless, if an evaluation tree intended for parallel evaluation has significantly higher maximum dimension than the best sequential evaluation tree, we are unlikely to obtain good speedup. From the distribution and return communication cost models in section 3 we can see that communication cost is mainly determined by the $max(d|, d2, r)$. We call this parameter $md$ and report it in tables 2, 3 and 4. From these tables we can see the larger the $md$ value, the higher the communication cost. Notice that the $md$ value for table 2 is always equal to $max\_dim - 1$. This artifact of the way SPI constructs its evaluation trees explain why the communication cost for SPI is always high. Available speedup under the BCA model is essentially exponential in $max\_dim - md$, since this difference reflect the extra computational burden we can reduce through parallelization.

It seems reasonable to assume that larger nets ought to be more "parallelizable." We can roughly measure the quality of a parallel factoring, then, by taking the ratio $(max\_dim - md)/max\_dim$. Given a $max\_dim$, the higher this ratio, the better the speedup. We report this value in column dd, in each of the basic result tables, 2, 3, and 4. The dd values for the nets with small speedup, like nets 2, 5 and 8, shows that the low speedup obtained can be attributed to the smallness of the problem rather than the quality of the factoring.

**Communication/Memory tradeoffs**  The differences between the BCA-CP model and the dist-net model are: (1) there is no data distribution needed in the dist-net model; and (2) the dist-net model must store the entire network at each processor. This gives rise to two questions: (1) is there significantly more speedup available under the dist-net model; (2) does the BCA model offer a significantly lower per-processor memory requirement. Our experimental results in table 5 show that the communication cost for the dist-net model is about half of that for the BCA model. The real impact of communication cost, however, depends on the ratio of computation cost and communication cost. For those queries in which computation time is about 2 to 3 percent of communication time, see table 2, the speedup for dist-net model eval-uation would be double that obtained using BCA-CP model evaluation. However, when the computation cost is close to the communication cost, see table 3 and table 4, the additional speedup using the dist-net model is negligible.

---
[6]In doing this we relax our limitation on number of processors.

| # | BCA-cm | Dist-cm | BCA-mem | Dist-mem | memory | mem/Dist-mem |
|---|--------|---------|---------|----------|--------|--------------|
| 1 | 3.11+7 | 1.60+7 | 1.57+6 | 2.54+7 | 1.84+4 | 1380 |
| 2 | 1.48+6 | 7.39+5 | 6.15+3 | 1.31+6 | 1.28+3 | 1024 |
| 3 | 1.7+9 | 8.86+8 | 2.10+6 | 1.75+9 | 1.70+6 | 1029 |
| 4 | 3.16+5 | 1.56+5 | 4.10+3 | 2.62+5 | 2.56+2 | 1023 |
| 5 | 5.98+9 | 2.99+9 | 1.68+7 | 5.40+9 | 5.28+6 | 1022 |
| 6 | 1.51+11 | 7.63+10 | 3.22+9 | 1.50+11 | 1.47+8 | 1020 |
| 7 | 1.78+8 | 9.08+7 | 6.29+6 | 1.55+8 | 1.48+5 | 1047 |
| 8 | 3.64+5 | 1.81+5 | 4.10+3 | 2.70+5 | 2.64+2 | 1023 |

Table 5: Comparison between the dist-net model and the BCA-CP model.

| net | para-cp | %-cp | lp-cp | lp-speedup | lp-%-cp | %-time |
|-----|---------|------|-------|------------|---------|--------|
| 1 | 21 | 0.41 | 7 | 118.1 | 0.14 | 0.945 |
| 2 | 13 | 0.19 | 11 | 66.8 | 0.23 | 0.800 |
| 3 | 17 | 0.20 | 7 | 489.3 | 0.11 | 0.975 |
| 4 | 12 | 0.24 | 7 | 60.9 | 0.17 | 0.652 |
| 5 | 26 | 0.52 | 9 | 75.3 | 0.18 | 0.964 |
| 6 | 32 | 0.51 | 10 | 536.6 | 0.16 | 0.985 |
| 7 | 23 | 0.42 | 8 | 150.9 | 0.14 | 0.947 |
| 8 | 13 | 0.56 | 10 | 93.2 | 0.21 | 0.710 |

Table 6: Evaluation-tree Parallelism: the set-factoring(c) algorithm.

The memory requirements of the alternate evaluation models, in contrast, are quite dramatically different, see table 5. We calculated the memory requirement in the dist-net model as $max_{CP_s}(2^{d1} + 2^{d2} + 2^r)$. This value is an estimated optimal result because we ignore all but the largest conformal product in each query. The memory requirement for the BCA model for each processor[7] is calculated as the size of the total data sent to and from a single processor. This value is an upper bound because it assumes each processor must hold all intermediate results. Since the splitting strategy used in the BCA model often limits the parallelism for the last conformal product, which results in an artificially high memory use, we list the memory size ignoring this last conformal product in the column memory in table 5. Comparing the values in BCA mem and in memory, we can see that the memory used in the dist-model is about 1000 times higher than that used by the BCA model. Since we performed all these measurements assuming 1024 available processors, we can therefore conclude that the BCA model is effective in distributing the memory requirement as well as the computational burden.

---

[7]There is one processor which has the same size as in the dist-net model, at which results are aggregated. We ignore this processor in this section.

**Why does set factoring perform so well?** We did not initially expect that evaluation trees produced by the sequential version of set-factoring would perform so well. Why is set-factoring so effective in reducing communication cost? Checking the *md* column in the table 3, we can see that the values are not 1s. This means that, in contrast to SPI, set-factoring tends to construct evaluation trees in which the distributions being combined each contain many variables not in the other. Intuitively, we can understand this as a result of the "procrastination" inherent in set-factoring's greedy heuristic. Since set-factoring always seeks the minimum-cost conformal product it can perform next, it tends to produce bushy and balanced trees. But these same trees are exactly the type likely to exhibit large *md* values, precisely what we need for good speedup.

**Why doesn't set-factoring(c) perform even better?** We thought, at first, that considering communication cost in the set-factoring algorithm would reduce communication cost significantly so that the speedup would be further improved. Comparing the communication cost in the table 3 and the table 4, we can find that there is, in fact, little if any improvement. We believe that this indicates that set-factoring without communication cost has already found most of the available parallelism, or at least most of that which can be found through simple hill-climbing.

**Evaluation tree parallelism** The same arguments as above would lead one to expect that set-factoring should find more evaluation tree parallelism. That is, shallower, bushier trees should exhibit more evaluation tree parallelism. While this seems to be true, column %-time in table 6 indicates that even set-factoring finds little, if any, evaluation tree parallelism. The basic reason for this is that one conformal product typically dominates the total computation. In net 3, for example, out of 64 conformal products there is only one with dimension 35, one with 32 and six between 20 to 28. Largely as a result of this, the sum of computation costs in the longest path in any evaluation tree dominates the total computation cost. At least for these nets and this approach to inference, evaluation tree parallelism seems insignificant. If this result holds for more general classes of networks and extends to clustering approaches to belief-net inference [9], [7], as we suspect it does, then early claims that clustering-style inference techniques allow "distributed" revision of beliefs must be re-examined. We still consider it an open question, however, whether the dominance of a single or small number of large conformal products is a characteristic of a probability computation in general or an artifact of our approaches to evaluation tree construction.

**Other factors affecting speedup** There are two other factors that affect available speedup. Sometimes a computation consists of many small conformal products which cannot be parallelized. From the column para-CP in table 6 we can see that on average parallelizable conformal products are less than half of the total. This usually affects only smaller queries, however. More critical, we believe, is the strategy used for choosing splitting variables. It is possible, as mentioned earlier, to split of variables not in the result. When the input distributions are large and the result only contains a few variables, our choice to restrict splitting vars to those in the result can significantly limit speedup. One example of this is when the final conformal product has large input distributions, but only one result variable (the query variable). In such cases, better speedup would have been obtained had we split on variables not in the result. This would, however, have required that we include the cost of aggregating the result (which can be done in log time). However, the memory requirements results indicate this may be worthwhile.

## 7  Summary

We have presented the results of an experimental exploration of the feasibility of parallel computation of queries in large multiply-connected belief net. We presented two parallel architectures (BCA and Distnet) and three factoring strategies (SPI, set-factoring, and set-factoring(c)). Our results indicate that good speedups are available on current and expected hypercube architecture machines for multiply-connected networks, under reasonable assumptions about number of processors and task grainsize. Our results further indicate that this parallelism is available, not at the topological level, but only through parallelizing individual conformal product operations.